\documentclass[lettersize,journal]{IEEEtran}
\usepackage{amsmath,amsfonts}
\usepackage{array}
\usepackage[caption=false,font=normalsize,labelfont=sf,textfont=sf]{subfig}
\usepackage{textcomp}
\usepackage{stfloats}
\usepackage{url}
\usepackage{verbatim}
\usepackage{graphicx}
\usepackage{cite}
\usepackage[ruled]{algorithm2e}
\usepackage{amssymb}

\usepackage[pagebackref,breaklinks,colorlinks]{hyperref}

\usepackage[capitalize]{cleveref}
\crefname{section}{Sec.}{Secs.}
\Crefname{section}{Section}{Sections}
\Crefname{table}{Table}{Tables}
\crefname{table}{Tab.}{Tabs.}

\usepackage{booktabs}
\usepackage{pifont}
\usepackage{multirow}
\usepackage{bm}
\usepackage[table,xcdraw]{xcolor}

\begin{document}
\title{CRS-Diff: Controllable Remote Sensing Image Generation with Diffusion Model}

\author{Datao Tang, Xiangyong Cao, Xingsong Hou, Zhongyuan Jiang, Junmin Liu, Deyu Meng
\thanks{This work was supported in part by the National Key
Research and Development Program of China under Grant 2021ZD0112902 and in part by the China NSFC Projects under Contract 62272375 and Contract 12226004. (\emph{Corresponding author: Xiangyong Cao.})}
\thanks{Datao Tang and Xiangyong Cao are with the School of Computer Science and Technology and the Ministry of Education Key Lab for Intelligent Networks and Network Security, Xi’an Jiaotong University, Xi’an 710049, China (Email: caoxiangyong@xjtu.edu.cn).}
\thanks{Xingsong Hou is with the School of Information and Communications Engineering, Xi’an Jiaotong University, Xi’an, Shaanxi 710049, China.}
\thanks{Zhongyuan Jiang is with the School of Cyber Engineering, Xidian University, Xi’an, Shaanxi 710049, China.}
\thanks{Junmin Liu and Deyu Meng are with the School of Mathematics and Statistics and the Ministry of Education Key Laboratory of Intelligent Networks and Network Security, Xi’an Jiaotong University, Xi’an, Shaanxi 710049, China, and also with Pazhou Laboratory (Huangpu), Guangzhou, Guangdong 510555, China.}
}


\maketitle

\begin{abstract}
The emergence of generative models has revolutionized the field of remote sensing (RS) image generation. Despite generating high-quality images, existing methods are limited in relying mainly on text control conditions, and thus do not always generate images accurately and stably. In this paper, we propose CRS-Diff, a new RS generative framework specifically tailored for RS image generation, leveraging the inherent advantages of diffusion models while integrating more advanced control mechanisms. Specifically, CRS-Diff can simultaneously support text-condition, metadata-condition, and image-condition control inputs, thus enabling more precise control to refine the generation process. To effectively integrate multiple condition control information, we introduce a new conditional control mechanism to achieve multi-scale feature fusion, thus enhancing the guiding effect of control conditions. To our knowledge, CRS-Diff is the first multiple-condition controllable RS generative model. Experimental results in single-condition and multiple-condition cases have demonstrated the superior ability of our CRS-Diff to generate RS images both quantitatively and qualitatively compared with previous methods. Additionally, our CRS-Diff can serve as a data engine that generates high-quality training data for downstream tasks, e.g., road extraction. The code is available at \href{https://github.com/Sonettoo/CRS-Diff}{https://github.com/Sonettoo/CRS-Diff}.
\end{abstract}

 
	
\IEEEpeerreviewmaketitle

\begin{figure}
    \centering
    \includegraphics[width=0.9\linewidth]{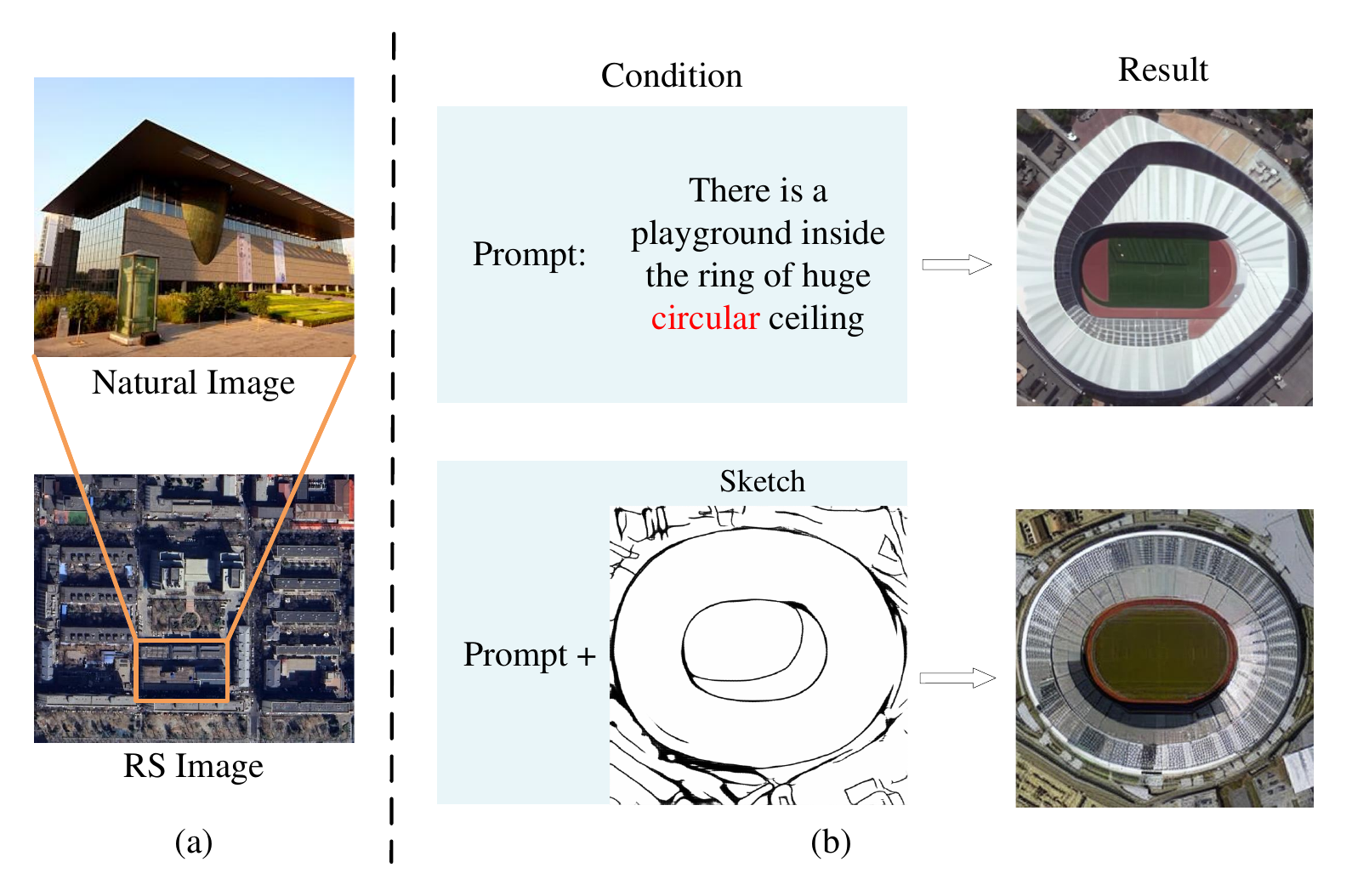}
    \caption{(a) Comparison between natural image and remote sensing (RS) image. The image content is the Capital Museum of China, sourced from Google Maps and Google Street View, respectively. As can be seen, RS imagery differs significantly from traditional RGB imagery in resolution, coverage area, and information richness. (b) Comparison of the generation results between the two control modes. The upper image is the generation result guided solely by text, while the lower image is the result guided by both text and sketch. As can be seen, the single text control condition fails to generate accurate image content while "text + sketch" conditions can succeed.}
    \label{fig:Comparison}
\end{figure}

\begin{figure*}
    \centering
    \includegraphics[width=0.9\linewidth]{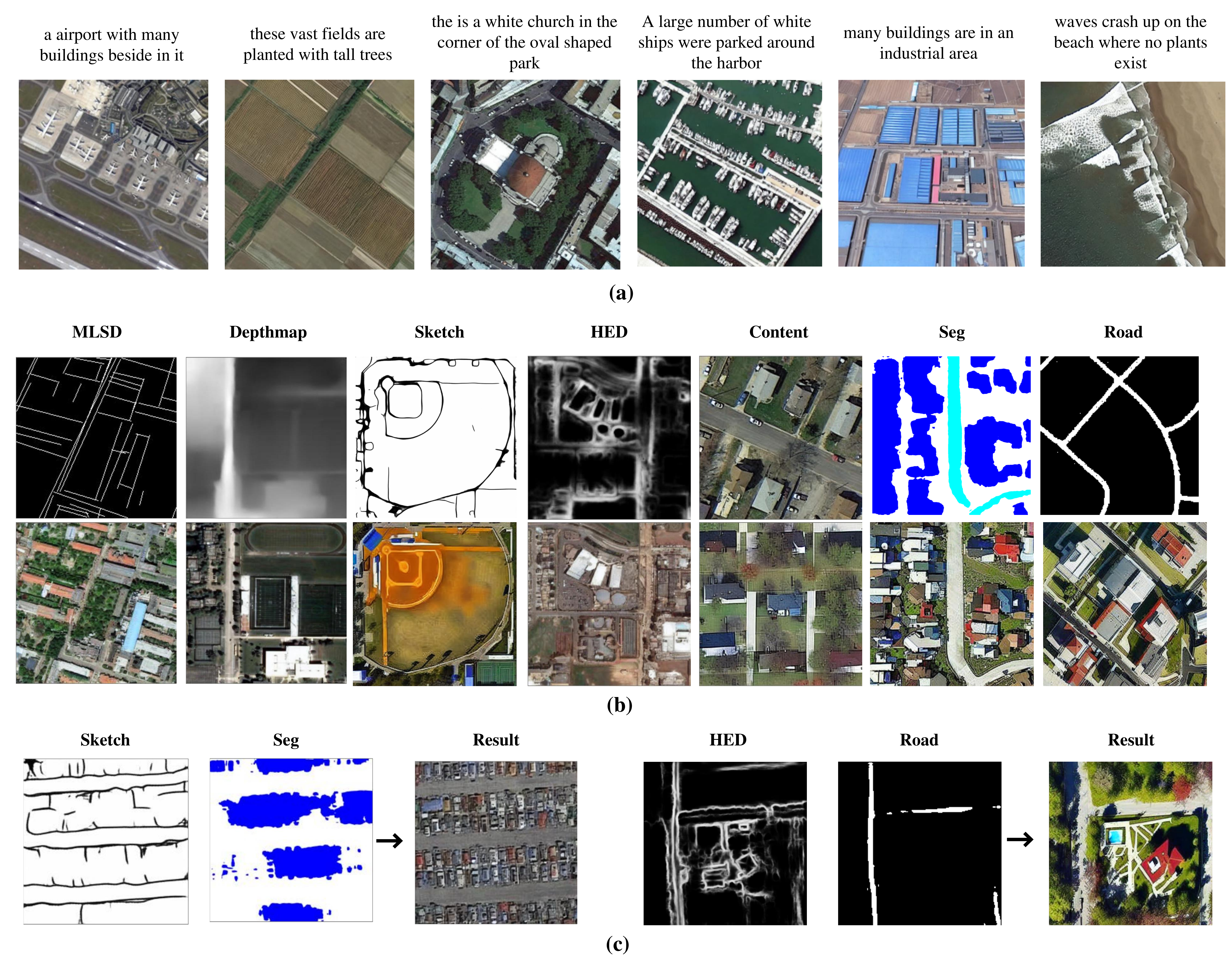}
    \caption{Visualisation results of our proposed CRS-Diff. (a) Singe text condition generation: the RS images are generated based only on text. (b) Single image condition generation: the RS images are generated based on the image condition. (c) Multi-condition image generation: the RS images are generated under the control of multiple conditions.}
    \label{fig:txt2img}
\end{figure*}

\section{Introduction}
\label{sec:intro}
\IEEEPARstart{D}{iffusion} models \cite{ho2020denoising,song2020score} are a class of probabilistic generative models that turn noise into a representative data sample. Recently, image generation based on diffusion model \cite{nichol2021glide,Feng_2023_CVPR,ramesh2022hierarchical,rombach2022high,croitoru2023diffusion,gu2022vector,song2020denoising,podell2023sdxl} has emerged as a hot research topic since the generated images exhibit high quality, e.g., generating realistic images \cite{nichol2021glide,Feng_2023_CVPR,ramesh2022hierarchical,rombach2022high}, transforming art styles \cite{10204740,zhang2023inversion}, image super-resolution \cite{wu2023hsrdiffhyperspectral,li2022srdiff,saharia2022image}, video generation \cite{ho2022imagen}, etc.
However, most existing diffusion models focus primarily on general image generation, with insufficient exploration in generating specific types of images, such as remote sensing (RS) images.

\begin{table}[h!]
    \caption{Comparison of existing controllable generative RS models.}
    \centering
    \scalebox{0.85}{
    \begin{tabular}{lccccc}
    \toprule
    Method &Fine-Tuning & Text & Metadata &Image  & Composable \\
      &  & Control & Control & Control & Control\\
    \midrule
    RSDiff \cite{sebaq2023rsdiff} & \ding{55} & \ding{51} & \ding{55} & \ding{55} & \ding{55} \\
    SatDM \cite{baghirli2023satdm} & \ding{55} & \ding{55} & \ding{55} & \ding{51} & \ding{55} \\
    RSFSG \cite{yuan2023efficient} & \ding{55} & \ding{55} & \ding{55} & \ding{51} & \ding{55} \\
    MetaEarth \cite{yu2024metaearth} & \ding{55} & \ding{51} & \ding{55} & \ding{51} & \ding{55} \\
    DiffusionSat \cite{khanna2023diffusionsat} & \ding{51} & \ding{51} & \ding{51} & \ding{51} & \ding{55} \\
    \midrule
    CRS-Diff (Ours) & \ding{51} & \ding{51} & \ding{51} & \ding{51} & \ding{51} \\
    \bottomrule
    \end{tabular}
    }
    \label{tab:comparison}
\end{table}

As shown in \cref{fig:Comparison} (a), the RS imagery differs significantly from the traditional RGB imagery in several ways, particularly in terms of resolution, coverage area, and information richness. The resolution of remotely sensed images is often very high to capture subtle surface features, unlike the standard resolution of traditional RGB images. In addition, RS images cover wide geographic areas and various environments, such as urban, rural, forest, and marine environments, providing extensive geographic information. In contrast, traditional RGB images usually capture only local areas, offering far less information richness than RS images. Therefore, the high resolution of RS images requires that generative models possess higher accuracy and detail capture capabilities, handle larger scale data, and maintain geographic information consistency during generation. Moreover, the rich information in RS images requires generative models to integrate and represent complex multidimensional data, as relying solely on simple textual control is often insufficient. As shown in the upper part of \cref{fig:Comparison} (b), textual descriptions, while providing some contextual information, are often inadequate for handling complex geographic and atmospheric data, making it difficult to accurately control the quality and content of the generated images. When generating scene details and regular buildings in the images, although the text-guided images exhibit largely similar features, they often contain distorted line segments and incomprehensible details that contradict the physical world. Therefore, we believe that more fine-grained conditional control is necessary to generate RS images. As shown in the lower part of \cref{fig:Comparison} (b), incorporating additional conditions (e.g., image sketch) into the image generation process enables the creation of more realistic images. Establishing this correspondence between conditions and images can expand the application scenarios of the generated model. Thus, additional control conditions need to be explored for RS image generation.



Currently, the research in RS image generation mainly includes GAN-based \cite{xu2018attngan,ruan2021dae,zhao2021text,tao2022df,zhou2022towards} and diffusion model-based approaches \cite{yuan2023efficient,sebaq2023rsdiff,liu2023diverse,khanna2023diffusionsat}. For example, Reed et al. developed StackGAN \cite{zhao2021text}, which employs stacked generators to produce clear RS images with the size of 256 $\times$ 256. However, GAN-based methods are unstable in the training process. In contrast, diffusion models exhibit superior generative ability and a relatively stable training process. As shown in \cref{tab:comparison}, there have been several controlled RS image generation models \cite{espinosa2023generate,yuan2023efficient,khanna2023diffusionsat,yu2024metaearth,baghirli2023satdm,sebaq2023rsdiff,toker2024satsynth}. For example, RSDiff \cite{sebaq2023rsdiff} proposes a novel cascade architecture for RS text-to-image generation using diffusion model. SatDM \cite{baghirli2023satdm} emphasizes the crucial role of semantic layouts in generating RS images and can produce RS images guided by semantic masks. Yuan et al. \cite{yuan2023efficient} notably generate high-quality RS images guided by semantic masks and introduce a lightweight diffusion model, obtained through a customized distillation process, to achieve fast convergence, addressing the inherent issue of prolonged training times in diffusion models. Recently, Yu et al. \cite{yu2024metaearth} proposed a guided self-cascading generation framework employing a novel noise sampling strategy, capable of generating images with diverse geographic resolutions across any region for downstream tasks. DiffusionSat \cite{khanna2023diffusionsat} incorporates the associated metadata such as geolocation as conditioning information to generate the RS image. However, these models lack control over the image detail level, still using text as the primary control condition and neglecting to incorporate image-related features as control signals. A single text-guided generated image can easily suffer from partial distortion (as shown in \cref{fig:Comparison} (b)), making it difficult to adapt to the high information density of RS images, rendering the generated images of limited use to downstream tasks.

To address these issues, in this paper, we propose CRS-Diff, i.e. a controllable remote sensing generation model. Specifically, a base diffusion model is first trained tailored for the RS domain based on the Stable Diffusion (SD) model \cite{rombach2022high}, which is capable of converting high-precision textual descriptions into RS images as shown in Fig. \ref{fig:txt2img} (a). Based on this, we integrated ControlNet \cite{zhang2023adding} to include two additional control signals in the diffusion model for the controlled generation of RS images. These two control signals adjust global and local condition information of the image, including six additional image control conditions (semantic segmentation mask, roadmap, sketch, etc.) and textual conditions (prompt, content image, and metadata encoding) as shown in Fig. \ref{fig:txt2img} (b). The optional combination of multiple conditions is controlled to ensure that the resulting RS images are visually realistic and accurately reflect specific geographic and temporal information. For the text condition, we concatenate directly with the original text encoding through an additional encoding step, leveraging the model's natural control mechanism. For the image condition, we explore multiscale feature fusion to coordinate different control conditions and efficiently implement the bootstrapping of generative process noise maps, making our method flexible enough to combine any conditions for image generation, as shown in Fig. \ref{fig:txt2img} (c). 

In summary, the contributions of our work are threefold: 
\begin{itemize}
\item We propose a new controllable RS generative model with diffusion models (CRS-Diff), which is a framework specifically designed for RS image generation. Different from previous RS generative models, our CRS-Diff can simultaneously support more types of controllable conditions, i.e., text, metadata and image.  

\item To effectively integrate multiple control information, we introduce a new conditional control mechanism to achieve multi-scale feature fusion to enhance the guiding effect of control conditions, thus broadening the image generation space. As far as we know, our CRS-Diff is the first multiple-condition controllable RS generative model, which is capable of generating high-quality RS images that meet specific requirements under the guidance of composite conditions. 

\item Experimental results have demonstrated the superiority of our proposed CRS-Diff in generating RS imagery that adheres to specific conditions and surpasses previous RS image generation methods both quantitatively and qualitatively. Additionally, our CRS-Diff can serve as a data engine that generates high-quality training data for downstream tasks, e.g., road extraction.
\end{itemize}

The rest of this paper is organized as follows. Section II provides a brief overview of related work. Section III details the technical aspects of the CRS-Diff implementation. Section IV describes the experimental design, presents the experimental results, and offers specific analyses. Finally, Section V presents the conclusions of the paper.

\section{Related work}

\subsection{Text-to-Image Generation}
Text-to-image generation that generates high-definition images corresponding to given textual descriptions has attracted significant attention in the multimodal field. Early research is primarily focused on GANs \cite{reed2016generative,reed2016learning,zhang2017stackgan}, with text-conditional GANs emerging as pioneering end-to-end differential architectures from character to pixel level. For example, Reed et al. \cite{reed2016learning} introduced the generative adversarial network, capable of generating 128$\times$ 128-pixel images. Additionally, Zhang et al. \cite{zhang2017stackgan} developed StackGAN, employing stacked generators to produce clear 256$\times$256 pixel RS images. However, these models face two main challenges: training instability and limited generalization to open-domain scenes. 

In addition to GAN-based methods, recent studies have shifted toward autoregressive models for text-to-image generation, using web-scale image-text pairs, such as DALL-E \cite{ramesh2022hierarchical}. These models demonstrate robust generative capabilities, particularly in zero-shot settings for open-domain scenes, starkly contrasting the small-scale data focus of GAN-based approaches. OpenAI's DALL-E, leveraging large transformer models and extensive training data, effectively maps language concepts to the pixel level, generating high-quality 256$\times$256 images. Furthermore, Yong et al. employed modern Hopfield layers \cite{demircigil2017model,schafl2022hopular} for hierarchical prototype learning \cite{liu2001evaluation} in text and image embeddings, aiming to extract the most representative prototypes and implement a coarse-to-fine learning strategy. These prototypes are then used to encapsulate more complex semantics in text-to-image tasks, enhancing the realism of generated RS images.

Diffusion Models \cite{ho2020denoising} are generative models that generate new images by gradually transforming an image from a Gaussian noise state to a target image. This process contains two main steps: the forward diffusion process and the reverse generation process. Compared to autoregressive models, diffusion models excel in generating more realistic images through a gradual denoising process. Numerous studies have since focused on enhancing the diffusion model, 
DALLE-2 \cite{ramesh2022hierarchical} enhances textual guidance capabilities through integration with the CLIP model, while GLIDE \cite{nichol2021glide} explores diverse guidance methodologies. Conversely, Stable Diffusion (SD) \cite{rombach2022high} augments the training data by leveraging the capabilities of the diffusion model, thus improving the generation results.

\subsection{Controlled diffusion models}
Controlled Diffusion Models (CDM) \cite{zhang2023adding,huang2023composer,mou2023t2i,khanna2023diffusionsat,zhao2024uni,li2023gligen} for text-to-image (T2I) generation aims to enable users to precisely dictate the content of generated images. Although traditional T2I models can generate images from text descriptions, users often experience limited control over the final output. Controlled diffusion models enable users to specify additional generative details, including style, color, and object positioning, through the introduction of enhanced controllable parameters or mechanisms. For example, ControlNet \cite{zhang2023adding}, GLIGEN \cite{li2023gligen} and T2I-Adapter \cite{mou2023t2i} incorporate additional control networks or control signals on top of the weights of SD to enable integrated control of multiple conditions to reduce training costs. The composer trains a large diffusion model from scratch through a new generative paradigm that allows flexibility in the construction of generative conditions, improves controllability, and achieves better results.

In particular, ControlNet \cite{zhang2023adding} can be used in conjunction with diffusion models such as Denoising Diffusion Probabilistic Models (DDPM) to augment the controllability and diversity of the generated images. By introducing additional control signals or conditions, such as textual descriptions, image attribute labels, etc., Zhao et al. \cite{zhao2024uni} proposed Uni-ControlNet, which supports various additional control signals or combinations of conditions. By fine-tuning the adapters while keeping the original SD model unchanged, Uni-ControlNet significantly reduces the training cost, necessitating only two additional adapters for effective outcomes. 

In the remote sensing field, many RS generative models have been proposed. For example, Espinosa et al. \cite{espinosa2023generate} proposed a pre-trained diffusion model that is conditioned on cartographic data to generate realistic satellite images. RSDiff \cite{sebaq2023rsdiff} introduces a new architecture consisting of two cascading diffusion models for RS text-to-image generation. Yuan et al. \cite{yuan2023efficient} introduced a lightweight diffusion model obtained through a customized distillation process, which enhances the quality of image generation via a multi-frequency extraction module and achieves fast convergence by resizing the image at different stages of the diffusion process. SatSynth \cite{toker2024satsynth} can simultaneously generate images and corresponding masks for satellite image segmentation, which can then be applied to data augmentation. DiffusionSat \cite{khanna2023diffusionsat}, has demonstrated the capability to generate high-resolution satellite data utilizing numerical metadata and textual captions. In contrast, we employed a sophisticated training strategy to design and introduce an additional control network, enabling our model to achieve composite control generation under various conditions.

\begin{figure*}
	\centering
	\includegraphics[width=1\linewidth]{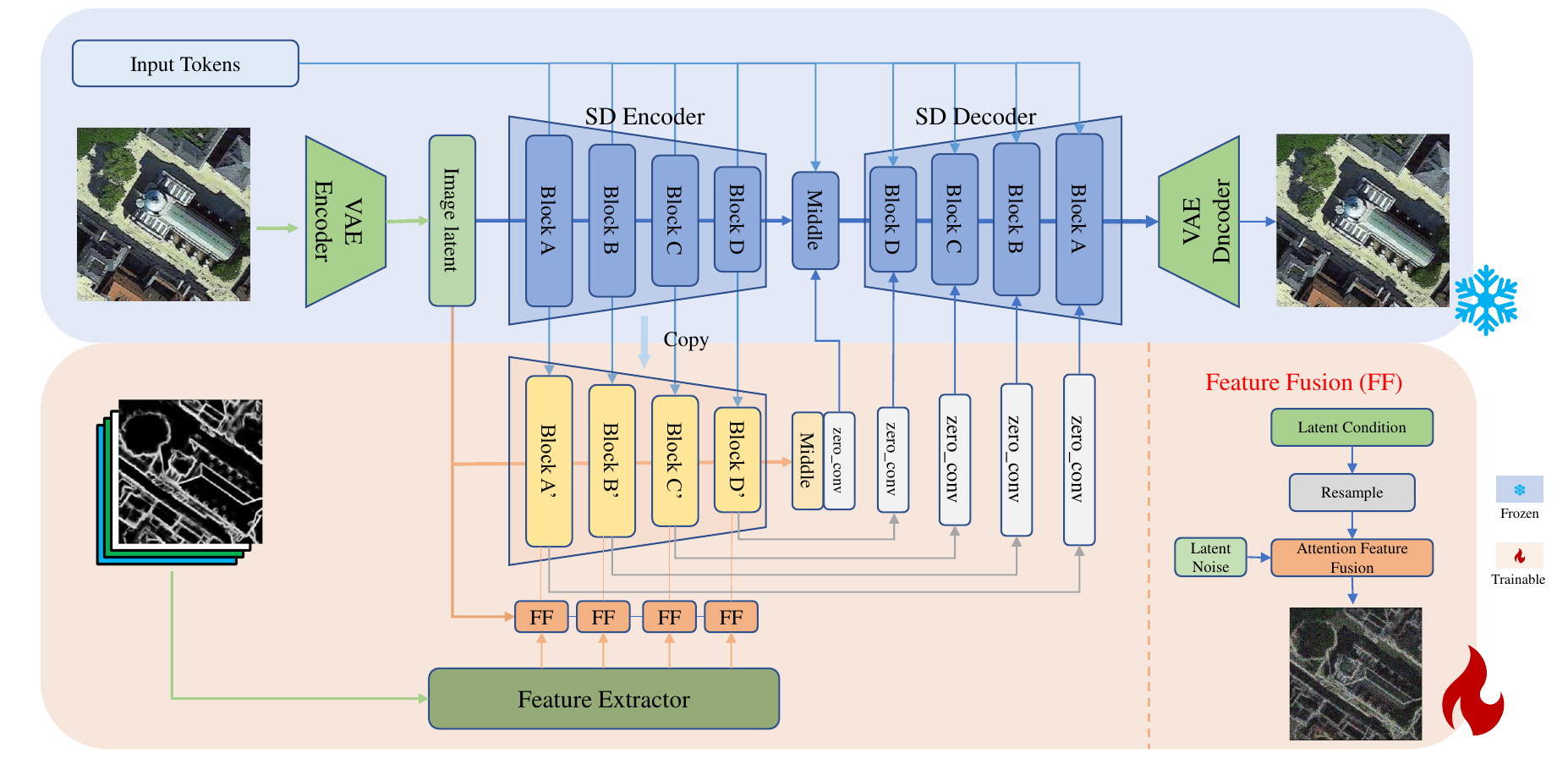}
	\caption{The overall architecture of our proposed CRS-Diff model. CRS-Diff is mainly based on Stable Diffusion (SD), that diffusion process is performed in latent space. The training of CRS-Diff contains two stages. In the first stage of the training process, we train the backbone U-Net network of SD on text-image pairs. The diffusion network obtained from this training is frozen (blue area) during the second training phase, and the encoder and intermediate blocks are copied into ControlNet to adapt to conditional inputs. In the second stage of training, we stack the conditional images as inputs and extract conditional features using a feature extractor. These features are gradually injected into the encoder of ControlNet (orange area) through a Feature Fusion (FF) module. Here, we use a convolutional network to reshape the obtained feature vectors to the current noise dimension and then integrate them with the noise output of the current block of the ControlNet encoder through Attention Feature Fusion (AFF), achieving multi-scale conditional injection.}
\label{fig:overall}
\end{figure*}

\section{Method}
\label{sec:method}
We introduce the diffusion model into the field of remote sensing image generation, aiming to enhance generic image generation capabilities for producing more realistic remote sensing images, subsequently introducing an optimized multi-conditional control mechanism that leverages text, image, and other multidimensional information to guide precise image generation and yield high-quality RS images. The construction of the model consists of two steps: initially, text-image pairs are utilized to train the generative diffusion model weights for RS images, building upon the traditional SD framework, and then the combination of multiple conditions (image conditions and text conditions) is implemented through a conditional control network. We will detail the two-stage training process for CRS-Diff, alongside illustrating the implementation of separate decompositions and combinations of training data.

\subsection{Text-to-Image generation}
Initially, we employed the Stable Diffusion (SD) \cite{rombach2022high} for text-to-image generation. This process involves utilizing a frozen variational autoencoder (VAE) encoder and decoder. The purpose is to convert each image \(x \in \mathbb{R}^{C \times H \times W}\) into its corresponding latent space variable \(z\), thus circumventing the direct learning of the original image's conditional data distribution given the text conditions \(p(x|\tau)\), and instead focuses on learning the feature distribution of the mapped image feature vector \(p(z|\tau)\). Text description corresponding to an image is encoded by a CLIP model~\cite{radford2021learning}, which subsequently guides the image generation during the denoising process via a cross-attention mechanism \cite{vaswani2017attention}. Thus, the training process of the diffusion model involves updating the latent space related to the U-Net. By predicting the noise added in the forward process and removing it in the reverse process, the model can learn the data distribution in the latent space. The training objective in this process is defined as follows:
\begin{equation}
   \min_{\theta} \mathcal{L}(\theta) = \mathbb{E}_{z, \epsilon, t} \left[ \|\epsilon - \epsilon_\theta(z_t, t, c)\|_2^2 \right],
\end{equation}
where \(\theta\) is the parameters of the model being optimized, \(\epsilon\) is the noise added in the forward process, \(\epsilon_\theta(z_t, t, c)\) denotes the prediction of the noise given the noisy data \(z_t\), time \(t\), and condition \(c\).

Additionally, a Classifier-free Guidance (CFG) \cite{ho2022classifier} mechanism is introduced:
\begin{equation}
   \hat{\epsilon}_{\theta}(z_t, c) = \omega \cdot {\epsilon}_{\theta}(z_t, c) + (1 - \omega) \cdot
 {\epsilon}_{\theta}(z_t),
\end{equation}
where \(z_t = a_t z_0 + \sigma_t\) and \(\omega\) denote the bootstrap weight. \(\hat{{\epsilon}_{\theta}}(z_t, c)\) is the output of the CFG mechanism. It is a weighted sum of the class-conditional output \({\epsilon}_{\theta}(z_t, c)\) and the unconditional output \({\epsilon}_{\theta}(z_t)\). 

\begin{figure}
	\centering
	\includegraphics[width=1\linewidth]{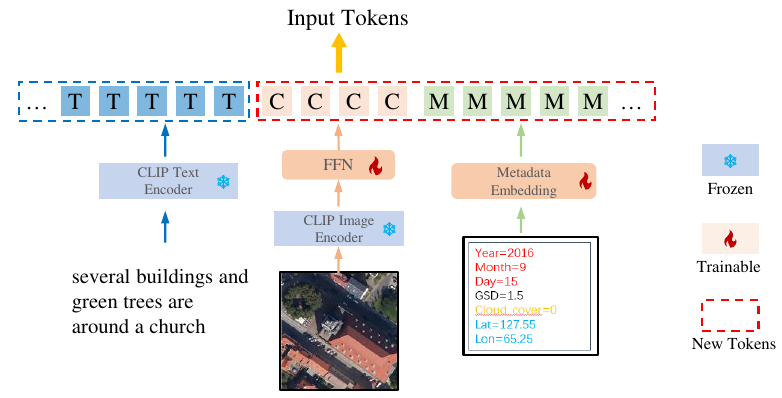}
	\caption{The raw text was encoded using a CLIP text encoder fine-tuned on RS images. The content image is initially encoded using the CLIP image encoder, and the resulting encoding is then converted into four additional text tokens by a Feed Forward Network(FFN). The metadata is first mapped into fixed intervals and then converted into the same number of tokens by an embedding layer. Finally, these processed encodings are concatenated, replacing the original text-encoded input.}\label{fig:model_1}
\end{figure}

Textual information serves as the sole guiding factor in this process. Simultaneously, we employ the pre-trained CLIP ViT-L-14 model \cite{rombach2022high}, fine-tuned on the RSICD RS image dataset, to amplify the effect of textual guidance. Under the original CFG framework, conditional noise prediction \({\epsilon}_{\theta}(z_t, c)\) is solely dependent on the model's processing of a given condition (e.g., text description). With the introduction of the CLIP bootstrap, this conditional prediction becomes further influenced by the similarity loss between the image and text as computed by the CLIP model. This implies that the calculation of \({\epsilon}_{\theta}(z_t, c)\) incorporates considerations for better aligning the resulting image with the textual description.

Our text-to-image generation methodology initially encodes an image \(x \in \mathbb{R}^{C \times H \times W}\) utilizing a static VAE, which is then converted into a latent representation \(z = E(x) \in \mathbb{R}^{C' \times H' \times W'}\). Subsequently, Gaussian noise is added to the latent image features to produce a noisy latent representation \(z_t = \alpha_t z + \sigma_t \varepsilon\), where \(\varepsilon\) represents the Gaussian noise component, and \(\alpha_t\) and \(\sigma_t\) are coefficients modulating the noise intensity.
The text caption \(\tau\) is encoded by a CLIP model \(T_{\theta}(\tau)\), generating the text embedding \(\tau'\). This is accomplished through a denoising model \(\epsilon_{\theta}(z_t; \tau', c)\). Finally, the denoised potential representation is up-sampled to the original image resolution using the VAE decoder, completing the image generation process. In this process, we use the Denoising Diffusion Implicit Models (DDIM) algorithm to sample, which can speed up the sampling process of diffusion models.

Additionally, the weights of the encoder \(E\), decoder \(D\), CLIP text encoder \(T_{\theta}\), and denoising model \(\epsilon_{\theta}\) are inherited from the SD 1.5 version. During the training process, only the denoising model \(\epsilon_{\theta}\) is updated to accelerate training convergence and exploit the rich semantic information in the SD model.

\subsection{Image Decoupling}
To address the issue of insufficient training data, the original image data is decoupled into the corresponding feature condition data using supplementary network structures, thus constructing a large-scale combined condition dataset that also functions as an interface during the inference process, and subsequently introducing nine conditions for formal model training.

\textbf{\textit{Caption}}: Under conditions where caption data is available, such as the RSICD dataset \cite{lu2017exploring}, we directly utilize the corresponding captions of remote sensing images. For other datasets, such as fMoW dataset \cite{christie2018functional}, we leverage the category information of remote sensing images, to synthesize captions. 

\textbf{\textit{HED (Holistically-nested Edge Detection)}}: A pre-trained deep neural network \cite{xie2015holistically} is employed to predict edges and object boundaries directly from the original image, thereby capturing high-level object boundary information and low-level details. 

\textbf{\textit{MLSD (Multiscale Line Segment Detection)}}: We use a pre-trained transformer-based model \cite{xu2021line} to detect straight line segments in remote sensing images.

\textbf{\textit{Depthmap}}: We use a pre-trained depth estimation model \cite{ranftl2020robust} to extract the Depthmap of the image, which approximates the layout of the image and aids in enhancing the model’s understanding of the remote sensing image's semantics.

\textbf{\textit{Sketch}}: Edge detection models \cite{simo2016learning} are applied to extract sketches from an image, focusing on the local details while conveying limited semantics.

\textbf{\textit{Road Map}}: In certain remote sensing images, greater emphasis is placed on road information, leading to the introduction of a pre-trained Separable Graph Convolutional Network (SGCN) \cite{zhou2021split} aimed at road extraction to yield single-channel road data.

\textbf{\textit{Segmentation Mask}}: we employ UNetFormer \cite{wang2022loveda}, a network specializing in remote sensing images, to extract semantic information and produce masks segmented into eight categories.

\textbf{\textit{Content}}: For content, the given image is considered as control information. Utilizing the Image Encoder in the pre-trained 
CLIP ViT-L-14 \cite{radford2021learning} model, the image is transformed into feature encoding, obtaining a global embedding. This approach offers more relevant guiding conditions than text descriptions alone.

\textbf{\textit{Metadata}}: In processing the RS image, it is crucial to incorporate additional metadata, such as temporal (year, month, day) and spatial (ground sampling distance, latitude, longitude, cloud cover) information. This information is first quantified and categorized, serving as input for the traditional diffusion model’s category guidance \cite{rombach2022high}, through the labeling of these categories. Additionally, these metadata are transformed into sequence tokens, which are incorporated into the text encoding as weak text control conditions.

\subsection{Multi-conditional fusion}
Based on the backbone structure outlined in the previous section, additional conditional control modules are added to reconstruct the original image from the decoupled image representation conditions. This process trains the model's multi-conditional generation capability. The known conditions are categorized into three types: image conditions, text conditions, and metadata. For each type of condition, we explore the corresponding condition injection methods and construct a feature extraction network that meets the requirements. The obtained condition features are then integrated with the ControlNet control strategy \cite{zhang2023adding} through feature fusion to achieve composite control of arbitrary conditions, as illustrated in \cref{fig:overall} and \cref{fig:model_1}.

\textit{Text Conditional Fusion:} We aim to establish a joint conditional bootstrapping mechanism that includes captions, content, and metadata, as illustrated in \cref{fig:model_1}. The text description, as the main bootstrap condition, is represented as \(y_{\text{t}}\) by obtaining a word embedding through a specialized CLIP text encoder. Different types of metadata are first mapped to values between 0 and 1 based on their value ranges, denoted as \(\mathbf{m} = [m_1, m_2, \ldots, m_n]\), where \(m_i\) denotes the \(i\)-th type of metadata. Subsequently, these normalized metadata values are encoded into vectors of uniform length using different Multi-Layer Perceptron (MLP) layers. These vectors are concatenated to form the metadata embedding \(y_{\text{m}}\) as
\begin{equation}
    y_{\text{m}} = [\text{MLP}_1(m_1); \text{MLP}_2(m_2); \ldots; \text{MLP}_n(m_n)],
\end{equation}
where \(\text{MLP}_i\) denotes the MLP used for the \(i\)-th metadata. 

Additionally, we introduce an image encoder to encode the content image and then use a Feed-Forward Network (FFN) to connect the feature vectors with the prompt encoding symbols, represented as:
\begin{equation}
    y_{}' = \text{Concat}\left(y_{\text{t}}, w_{\text{c}} \cdot \text{FFN}(y_{\text{c}}), w_{\text{m}} \cdot y_{\text{m}}\right),
\end{equation}
where \(w_c\) and \(w_m\) are the weights applied to the outputs of \(y_{\text{c}}\) and the \(y_{\text{m}}\), respectively. This network subsequently integrates these encodings with the caption's word embedding, replacing the original input tokens as the Key and Value in the cross-attention layer.

\textit{Image Conditional Fusion:} All decoupled image conditions, including Sketch, Segmentation mask, Depthmap, HED, Road map, and MLSD, are utilized as local control information. As shown in \cref{fig:overall}, this part of CRS-Diff is based on the ControlNet, with the SD weights initially fixed to replicate the encoders and the structure and weights of the intermediate blocks. It is worth noting that the feature extractor, consisting of stacked convolutional neural networks, contains multiple feature condition information in the feature map. We then fuse the latent features with the denoising latent variables through a feature fusion network. Consistent with Uni-ControlNet \cite{zhao2024uni}, we perform the feature injection step in four downsampled ResNet blocks within the U-Net structure of the diffusion model.

Specifically, given a set of image conditions \( \{c_i\}_{i=1}^n \), image conditional information processing can be represented as a function \( \mathcal{F} \), which maps the set of conditions to a feature space that equations with the input noise dimensions. This mapping is formally defined as \( \mathcal{F} : \{c_i\} \rightarrow \mathbb{R}^{c \times d \times d} \), where \( d \) represents the dimensionality of the latent feature map. The transformation leverages a series of convolutional and pooling layers to effectively capture the spatial hierarchies and semantic features of the control information, ensuring that the generated features are representative of the underlying conditions. Based on this, we resample the obtained feature map to the same dimension as the current latent variable and use Attention Feature Fusion (AFF) \cite{dai2021attentional} to achieve a better fusion of the noise and feature images. This process replaces the inputs of the original residual block of the U-Net, allowing us to achieve multi-conditional information injection, represented as:
\begin{equation}
    \mathbf{z}' = \text{AFF}(\mathbf{z}, \text{Resample}(\mathcal{F}(\{c_i\}), \text{dim}(\mathbf{z}))),
\end{equation}
where \(\mathbf{z}\) is the current latent variable, \(\{c_i\}\) is the set of image conditions, \(\mathcal{F}(\{c_i\})\) extracts the feature map from the conditions, and \(\text{Resample}(\cdot)\) adjusts the feature map dimensions to match \(\mathbf{z}\), \(\text{AFF}(\cdot)\) fuses the resampled feature map with \(\mathbf{z}\), \(\mathbf{z}'\) is the fused latent variable. This formula replaces the inputs of the original residual block of the U-Net, allowing us to achieve multi-conditional information injection.

\subsection{Training Strategy}
Our CRS-Diff employs a two-stage training strategy to train the native Stable Diffusion (SD) architecture and the ControlNet architecture within its framework, respectively. Initial training is conducted using SD 1.5 weights on a text-to-image RS dataset, aiming to develop a high-precision diffusion model for text-to-image generation that serves as the backbone of the ControlNet structure. This foundation enables the model to accurately guide the denoising process, leveraging a combination of multiple conditions through joint training. For both text and image control conditions, individual conditions are omitted with an independent probability of 0.5, and all conditions with a joint probability of 0.1, in accordance with Classifier-free Guidance. The dropout probability of certain conditions will be adjusted during experiments based on performance. Each condition element is treated as a distinct bootstrap condition, with single or multiple conditions being omitted at certain probabilities, enabling the model to learn a broader array of condition combinations.

\section{Experiments}
\label{sec:exper}
\subsection{Datasets}
During the training stage, our CRS-Diff used the following datasets:
\begin{itemize}
\item{RSICD dataset} \cite{lu2017exploring}: This dataset is designed specifically for image captioning in remote sensing imagery, and it contains 10,921 aerial remote sensing images accompanied by captions in natural language, sized at \(224 \times 224\). 

\item{fMoW dataset} \cite{christie2018functional}: This dataset is a large-scale remote sensing image dataset, and fMoW includes the spatio-temporal and category information for each image. Notably, only the RGB images, sized at \(224 \times 224\), are used, with relevant metadata extracted from a total of 110,000 images for training the multi-conditional control model.

\item{Million-AID dataset} \cite{long2021creating}: This dataset is a benchmark dataset for remote sensing scene classification, and contains millions of instances, featuring 51 scene categories with 2,000 to 45,000 images per category.
\end{itemize}

\subsection{Implementation details}
In the initial backbone model training phase, the following steps are taken: The RSICD dataset is fine-tuned over 10 epochs using the U-Net and the AdamW optimizer with a learning rate of \(1 \times 10^{-5}\). The input images are resized to \(512 \times 512\), with the model's parameters totaling approximately 0.9 billion. For the sampling process, DDIM is utilized, setting the number of time steps to 100 and the classifier-free bootstrap scale to 7.5. 

During the conditional control phase, original images from the fMoW and Million-AID datasets are organized into 200,000 text-image pairs. These images are then segmented into multiple conditional representations through annotator networks. This process involves randomly combining single or multiple conditional information, including road maps, MLSD, content, and extracted raw captions, metadata, and fine-tuning the conditional control network over 5 epochs. The AdamW optimizer, with a learning rate of \(1 \times 10^{-4}\), is used throughout this process, resizing both the input image and the local conditional graph to \(512 \times 512\). Experiments are conducted on NVIDIA GeForce RTX 4090 and NVIDIA RTX A100 GPUs, with a batch size of 8.

\begin{figure*}
    \centering
    \includegraphics[width=1\linewidth]{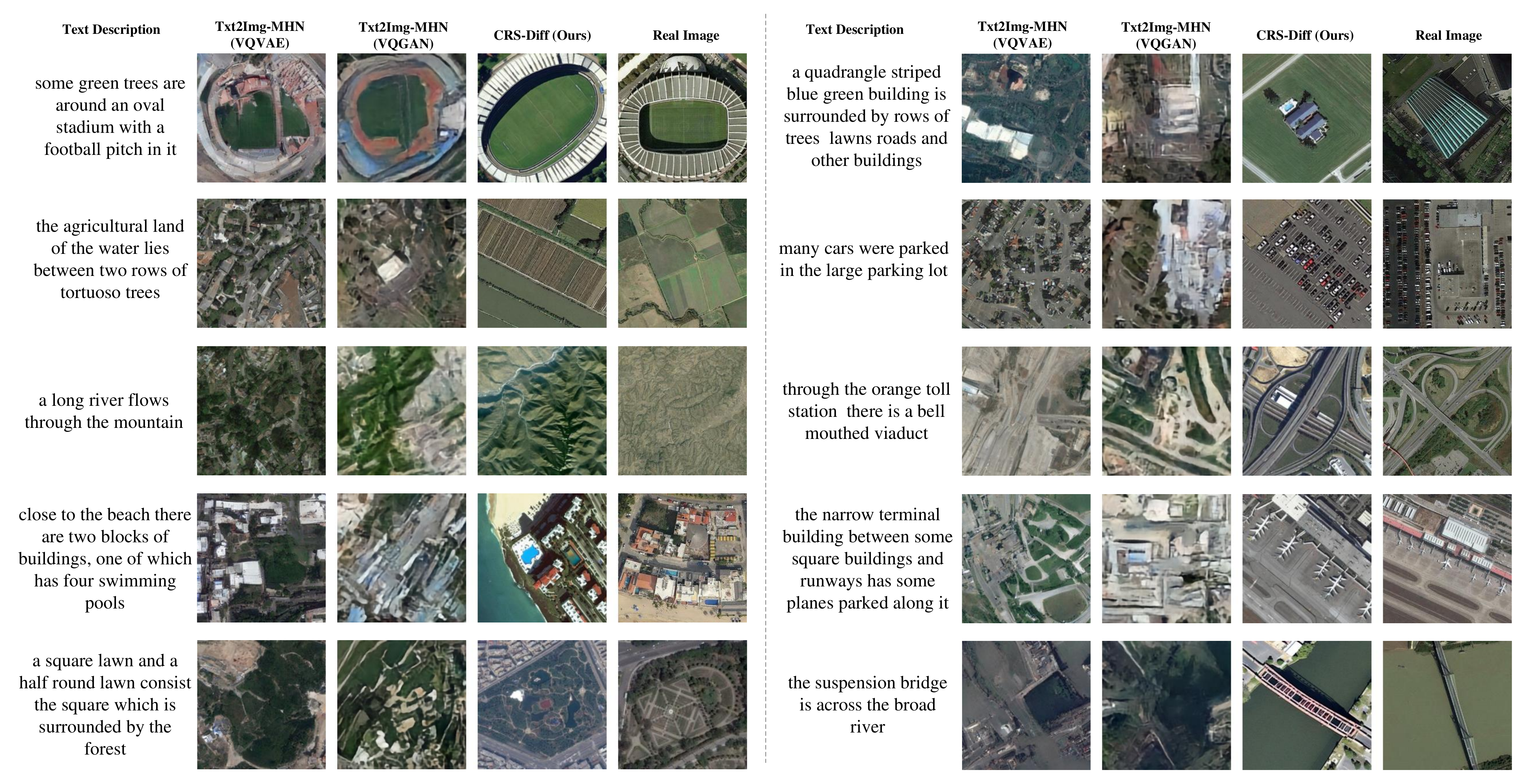}
    \caption{Visual comparison of different text-to-image generation methods based on text descriptions on the RSICD test set.}
    \label{fig:txt2img_2}
\end{figure*}

\begin{table*}
    \caption{Comparisons results between our CRS-Diff and other methods regarding four evaluation metrics (i.e., Inception score, FID score, CLIP score, and Zero-Shot classification OA on the RSICD test set). }
    \begin{center}
        {
        \scalebox{1}{
            \begin{tabular}{lcccc}
                \toprule
                Method & Inception Score $\uparrow$ & FID Score $\downarrow$ & CLIP Score ($\times$100) $\uparrow$& Zero-Shot classification OA $\uparrow$\\
                \midrule
                    Attn-GAN \cite{xu2018attngan}        & 11.71  & 95.81   & 20.19  & 32.56\\
                    DAE-GAN \cite{ruan2021dae}         & 7.71  & 93.15   & 19.69  & 29.74\\
                    StrucGAN \cite{zhao2021text}        & 5.84  & –     & –   & – \\
                    DF-GAN \cite{tao2022df}          & 9.51  & 109.41  & 19.76  & 51.99\\
                    Lafite \cite{zhou2022towards}          & 10.70  & 74.11   &\textbf{22.52}  & 49.37\\
                    \midrule
                    DALL-E \cite{ramesh2021zero}          & 2.59  & 191.93  & 20.13  & 28.59\\
                    Txt2Img-MHN (VQVAE) \cite{xu2023txt2img}    & 3.51  & 175.36  & 21.35  & 41.46\\
                    Txt2Img-MHN (VQGAN) \cite{xu2023txt2img}    & 5.99  & 102.44  & 20.27  & 65.72\\
                    \midrule
                    RSDiff \cite{sebaq2023rsdiff} & 7.22  & 66.49  & -  & -\\
		          SD 1.5 (original) \cite{rombach2022high}      & 6.33 & 167.19 & 19.92 & 32.81\\
                    SD 1.5 (fine-tune)     & 16.48 & 60.14 & 20.98 & 63.37\\
                    \midrule
                    CRS-Diff (Ours)            &\textbf{18.39} &\textbf {50.72}   & {20.33}  &\textbf {69.21}\\
                    \bottomrule
            \end{tabular}
            }
        }

    \end{center}
    \label{tab:result1}
    
\end{table*}

\subsection{Evaluation metrics}
For text-to-image generation tasks, we utilize four metrics to assess the effectiveness of image generation under the single text condition, namely the Inception Score (IS) \cite{salimans2016improved}, the Fréchet Inception Distance (FID) \cite{heusel2017gans}, the CLIP Score \cite{radford2021learning}, and the Overall Accuracy (OA) \cite{xu2023txt2img} for zero-shot classification. We evaluate the performance of our proposed method and the baselines under identical settings.

\begin{equation}
   IS = \exp \left( \mathbb{E}_{\mathbf{x} \sim p_g} \left[ D_{\text{KL}}(p(y|\mathbf{x}) || p(y)) \right] \right),
\end{equation}
where \( p(y|\mathbf{x}) \) is the conditional label distribution given an image \(\mathbf{x}\) and \( p(y) \) is the marginal label distribution. \( D_{\text{KL}} \) denotes the Kullback-Leibler divergence, and \( p_g \) represents the distribution of generated images.

\begin{equation}
   FID = ||\mu_r - \mu_g||^2 + \text{Tr}(\Sigma_r + \Sigma_g - 2(\Sigma_r \Sigma_g)^{\frac{1}{2}}),
\end{equation}
where \(\mu_r\) and \(\Sigma_r\) are the mean and covariance of the real images' features, and \(\mu_g\) and \(\Sigma_g\) are the mean and covariance of the generated images' features.

Notably, we utilize the zero-sample classification Overall Accuracy (OA) to evaluate the ability of the generative model to generalize across unseen categories. The computation of the OA for zero-sample classification is summarized in the following steps: 
\begin{enumerate}
\item Train a classifier (e.g., ResNet \cite{resnet}) by utilizing the images produced by the generative model as training data. 

\item Test the accuracy of this classifier in categorizing images within a set of real, yet unseen categories encountered during the training phase.
The formula is:
\end{enumerate}
\begin{equation}
    OA = \frac{1}{n} \sum_{i=1}^{n} \mathbf{1}(y_i = \hat{y}_i),
\end{equation}
where $y_i$ is the true label, $\hat{y}_i$ is the predicted label, $\mathbf{1}$ denotes the indicator function, and $n$ is the number of samples in the test set.

For conditional image generation tasks, we utilized three metrics (i.e., SSIM, mIoU, and CLIP score) to evaluate the performance of conditional image generation. We evaluated our proposed method and the baseline methods under identical settings. The calculation formulas for the SSIM and mIoU metrics are as follows:
\begin{equation}
\label{SSIM}
SSIM(x, y) = \frac{(2\mu_x \mu_y + c_1)(2\sigma_{xy} + c_2)}{(\mu_x^2 + \mu_y^2 + c_1)(\sigma_x^2 + \sigma_y^2 + c_2)},
\end{equation}
where \(\mu_x\) is the mean of image \(x\), \(\mu_y\) is the mean of image \(y\), \(\sigma_x^2\) is the variance of image \(x\), \(\sigma_y^2\) is the variance of image \(y\), \(\sigma_{xy}\) is the covariance of images \(x\) and \(y\), and \(c_1\) and \(c_2\) are constants to stabilize the division.

\begin{equation}
\label{mIoU}
mIoU = \frac{1}{n} \sum_{i=1}^{n} \frac{|P_i \cap G_i|}{|P_i \cup G_i|},
\end{equation}
where \(n\) is the number of classes, \(P_i\) is the predicted region for class \(i\), and \(G_i\) is the ground truth region for class \(i\).

\subsection{Comparison and Analysis}

\subsubsection{\textbf{Text-to-Image generation}} We trained CRS-Diff on the RSICD dataset using solely text as the initial input and compared it with recent state-of-the-art (SOTA) methods. 

\textit{Qualitative analysis.} In \cref{fig:txt2img_2}, generated images using various methods, such as CRS-Diff, Txt2Img-MHN (including VQVAE and VQGAN) \cite{xu2023txt2img}, are illustrated. Notably, our proposed CRS-Diff demonstrates the capability to generate clearer and more realistic images compared with other methods. For instance, when confronted with complex textual descriptions, exemplified by the phrase on the left side of the fifth line: ``a square lawn and a half round lawn consist the square which is surrounded by the forest'', the model proficiently deciphers the semantic content. Moreover, the model adeptly identifies the semantic information correlated with the textual descriptions, accurately reflecting this in the results. It also excels at synthesizing more appropriate images in response to shape descriptions like ``square'' and ``long.'' Furthermore, CRS-Diff distinctly grasps the concept of quantity, a feature intuitively evident in its handling of numerical descriptors like ``some'', ``many'', or ``two'' (as seen on the left side of the first line). Besides, our model can accurately simulate real lighting conditions, and coordinate elements such as color and texture.

\begin{figure}
	\centering
	\includegraphics[width=0.85\linewidth]{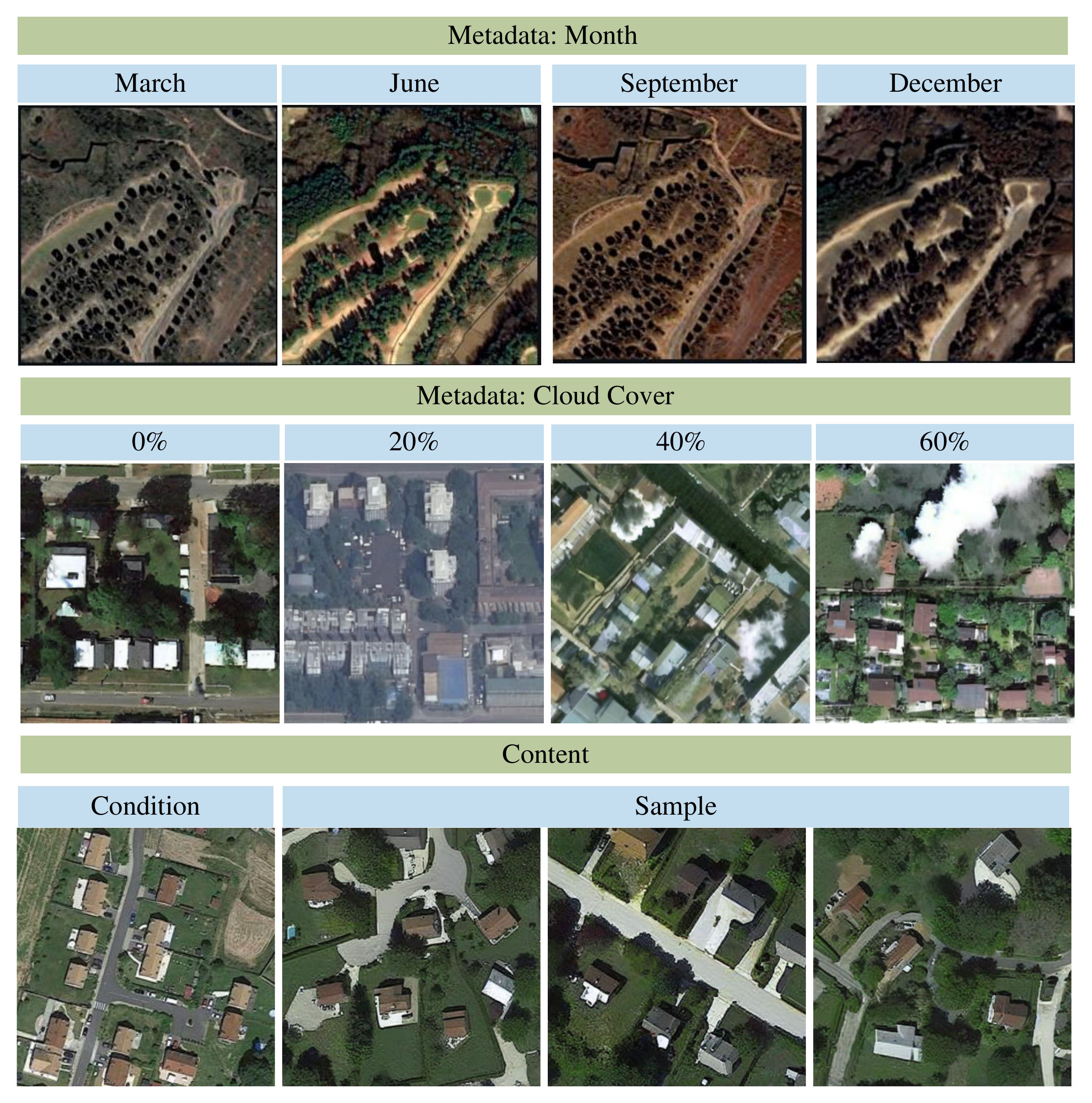}
	\caption{Visual comparison results of generated RS images from metadata (i.e., month and cloud cover) and image content. Except for the image content condition, metadata is used together with the textual descriptions.}\label{fig:control_image10}
\end{figure}

\textit{Quantitative analysis.} We compared CRS-Diff with Attn-GAN \cite{xu2018attngan}, DAE-GAN \cite{ruan2021dae}, StrucGAN \cite{zhao2021text}, DF-GAN \cite{tao2022df}, Lafite \cite{zhou2022towards}, DALL-E \cite{ramesh2021zero}, Txt2Img-MHN (including VQVAE and VQGAN) \cite{xu2023txt2img}, RSDiff \cite{sebaq2023rsdiff} and SD (fine-tuned on sd1.5) \cite{rombach2022high}. The generated results are quantitatively analysed using the RSICD test set, employing four evaluation metrics: zero-shot classification OA, initial score, CLIP score, and FID score, as delineated in \cref{tab:result1}. Our method surpassed the baseline in three metrics and achieved second place in the Inception Score. Intriguingly, the performance in the CLIP score is not as high as anticipated, possibly due to the specificities of the CLIP model employed in our evaluation. Nevertheless, the proposed CRS-Diff demonstrates excellent performance in controllability and generation quality, meeting the demands of practical applications like urban planning and laying a foundation for future research on controllable generation.

\subsubsection{\textbf{Single-condition image generation}} Except for the text condition, CRS-Diff can support more conditions to guide the model towards generating more refined images.

\begin{table*}

\caption{Results comparison (i.e., SSIM, mIoU, and CLIP Score metrics.) between our CRS-Diff with ControlNet and Uni-ControlNet on the RSICD test set in terms of seven image-based conditions.}
\centering
\scalebox{0.9}{
    \begin{tabular}{lccccccc}
    \toprule[1pt]
   Method & HED & MLSD & DepthMap & Sketch & Seg & Road & Content \\
     & (SSIM) & (SSIM) & (SSIM) & (SSIM) & (mIoU) & (mIoU) & (CLIP Score) \\
    \midrule
    ControlNet \cite{zhang2023adding} & 0.4363 &\textbf{0.8345}  & 0.8003  & 0.2895  & \textbf{0.3841}  &0.3458  &0.7789  \\
    Uni-ControlNet \cite{zhao2024uni} & 0.4475 & 0.7311 & \textbf{0.8743} & 0.5288 & 0.3041 & 0.3351 & 0.8453 \\
    \midrule
    CRS-Diff (Ours) & \textbf{0.4548} & 0.7455 & 0.8705 & \textbf{0.5676} & 0.2982 & \textbf{0.3787} & \textbf{0.8620} \\
    \bottomrule[1pt]
    
    \end{tabular}
}
\label{tab:result2}
\end{table*}

\begin{table*}
\caption{Results comparison (i.e., FID metric.) between our CRS-Diff with ControlNet and Uni-ControlNet on the RSICD test set in terms of seven image-based conditions.}
\centering
\scalebox{1}{
    \begin{tabular}{lccccccc}
    \toprule[1pt]
    Method & HED & MLSD & DepthMap & Sketch & Seg & Road & Content \\
    \midrule
  ControlNet \cite{zhang2023adding} & 31.26 & 62.21 & 58.27 & 66.54 & 72.65 & 102.57 & 63.98 \\
Uni-ControlNet \cite{zhao2024uni} & 44.76 & 59.89 & \textbf{47.10} & \textbf{58.73} & 79.15 & \textbf{93.08} & 53.97 \\
    \midrule
CRS-Diff (Ours) & \textbf{30.18} & \textbf{55.75} & 54.40 & 68.29 & \textbf{70.05} & 94.16 & \textbf{44.93} \\
    \bottomrule[1pt]
    
    \end{tabular}
}
\label{tab:result3}
\end{table*}

\textit{Qualitative analysis. }\cref{fig:control_image10} shows the visual comparison results of generated RS images from a single metadata condition. To mitigate the risk of inaccurate results due to conflicts between the content condition and the semantics of the textual description, we added textual guidance to all conditions except the content condition. The content condition provides richer semantic information. Metadata control proved more challenging, so we chose salient attributes such as month and cloud cover to offer more granular control.

\begin{figure*}
	\centering
	\includegraphics[width=0.8\linewidth]{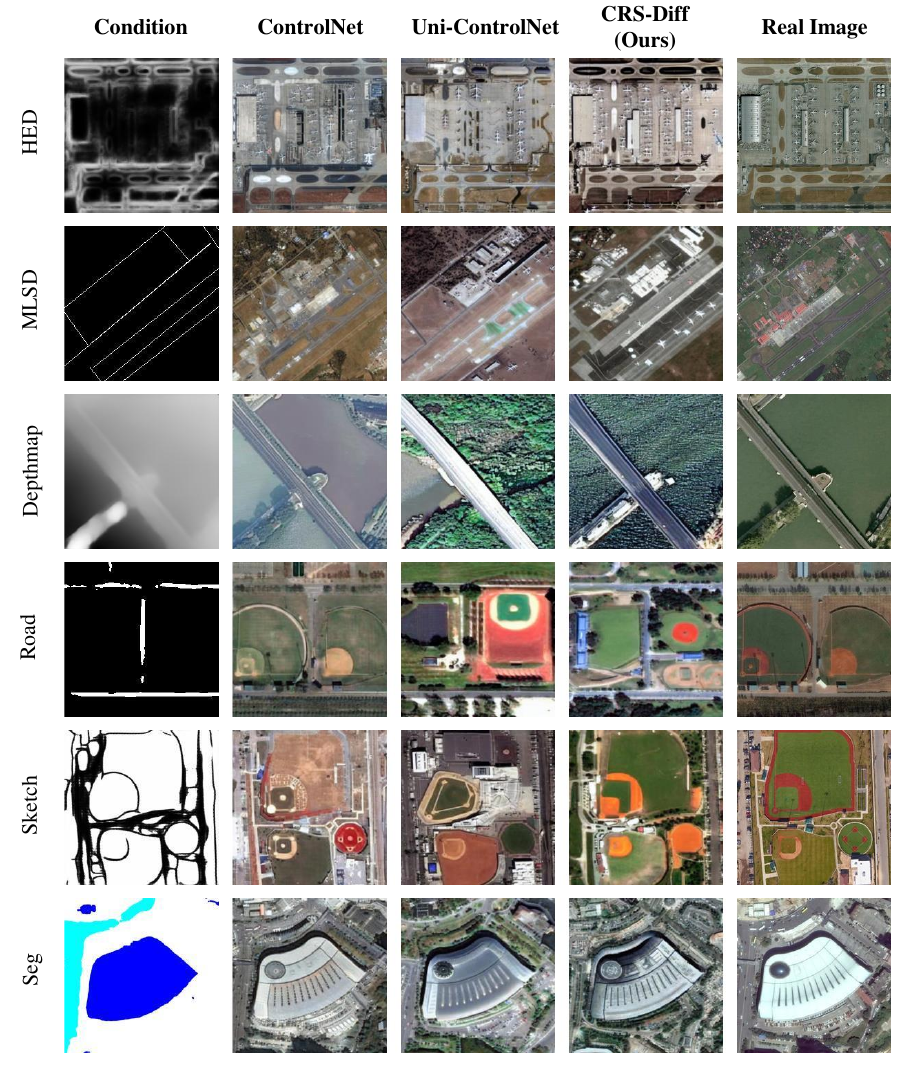}
	\caption{Visual comparison results of generated RS images from single image condition. All the conditions are used together with the textual descriptions.}\label{fig:control_image1}
\end{figure*}

\cref{fig:control_image1} shows the visual comparison results of generated RS images from a single image condition. In HED and Sketch, intuitive image control is achieved by restricting the boundary and contour information of the generated image. Features, e.g., segmentation masks and roadmaps, can provide richer semantic information, which can be efficiently interpreted by CRS-Diff to influence the generated outputs. Conversely, CRS-Diff generates images with clear texture details and coherent scene relationships, enabling better comprehension even in areas not covered by feature conditions.

\textit{Quantitative analysis}
We compare CRS-Diff with ControlNet and Uni-ControlNet for quantitative evaluation on a test set of RSICD at a resolution of \(512 \times 512\). We randomly select one caption per image from the test set to be used as textual bootstrap information, obtaining 1k pieces of generated images for the quality evaluation. We employ the image decoupling method mentioned earlier to obtain more control conditions for constructing conditional data. Single condition generation (in addition to metadata) is restricted for quantitative evaluation. For Uni-ControlNet and ControlNet (Multi-ControlNet), the same dataset is used to train the conditional generation capabilities of the seven conditions. For HED, MLSD, Sketch, and Depthmap, we compute the SSIM of the generated images corresponding to the decoupled conditions. For the semantic segmentation mask and road map, we compute the mIoU. For the Content condition, considered as a text markup, we compute the CLIP Score using the CLIP model fine-tuned to the remotely sensed images. The specific results are shown in \cref{tab:result2}. Our method achieves the best results on four metrics. Additionally, we calculate the FID metrics for the generated images, with specific results shown in \cref{tab:result3}. The experimental results demonstrate that CRS-Diff has excellent generative capabilities and quantitatively superior performance in most conditions compared to existing methods.

\begin{figure*}
	\centering
	\includegraphics[width=0.85\linewidth]{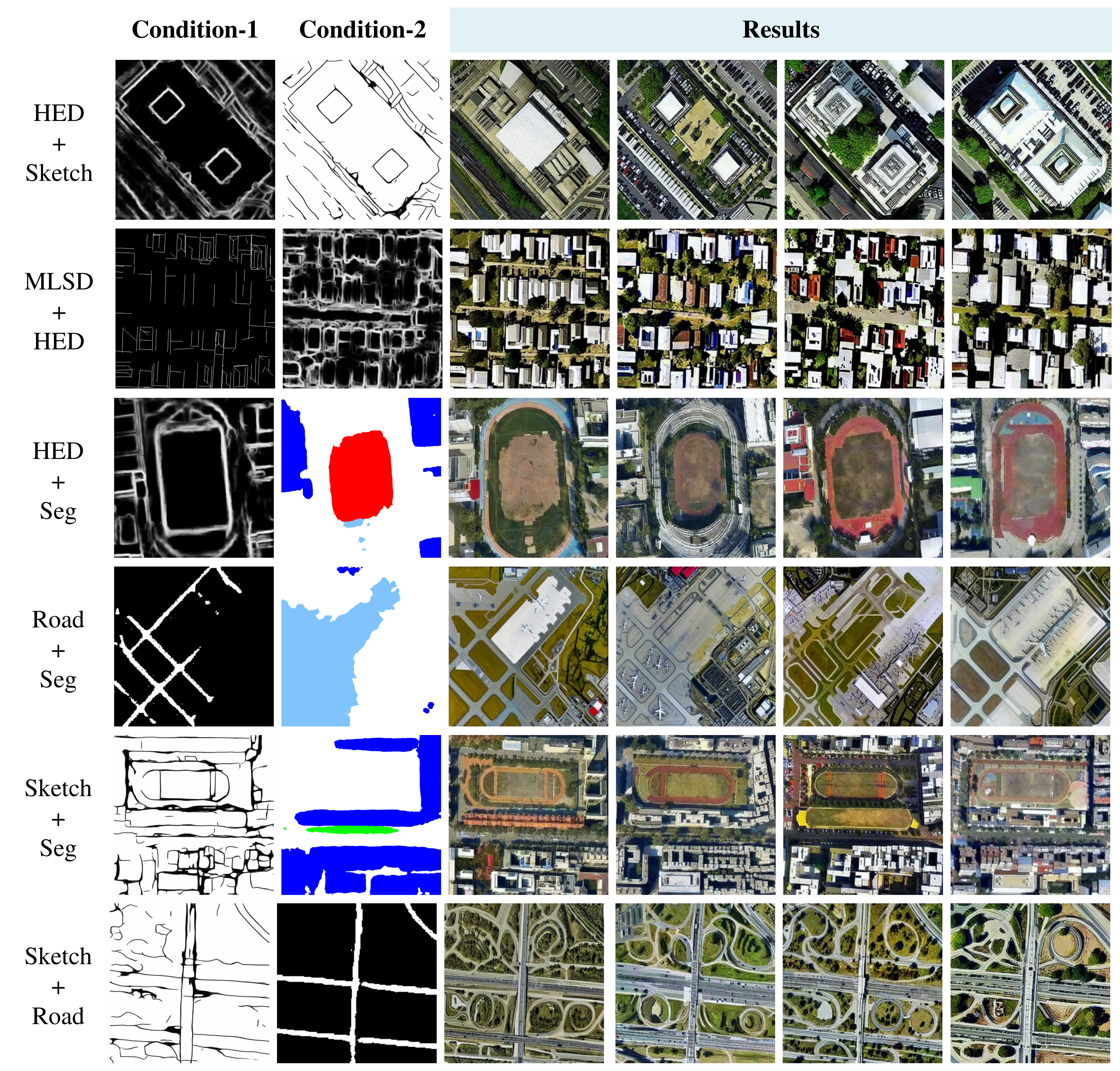}
	\caption{Visual comparison results of generated RS images under multiple condition control. Except for the content condition, all these conditions are used together with the textual descriptions. }\label{fig:control_image2}
\end{figure*}

\subsubsection{\textbf{Multi-condition image generation.}} We conducted tests under multiple conditions, but it is worth noting that conflicts may arise among these conditions, and in the context of RS images, even minor distortions can render the entire image meaningless, distinguishing from natural images. Therefore, in \cref{fig:control_image2}, we present the results generated under the influence of multiple conditions, striving to minimize conflicts between these conditions. Utilizing textual guidance, CRS-Diff excels in generative capacity, controllability, and realism, successfully completing the synthesis of the target image. Meanwhile, the generated images demonstrate sufficient diversity.

\begin{table}
\caption{Ablations of Backbone model (ReB) and Feature Fusion network (FF).}
\centering
\scalebox{0.75}{
    \begin{tabular}{ccccccccc}
    \toprule[1pt]
    ReB & FF & HED & MLSD & DepthMap & Sketch & Seg & Road & Content \\
     &  & (SSIM) & (SSIM) & (SSIM) & (SSIM) & (mIoU) & (mIoU) & (CLIP Score) \\
    \midrule
     &  & 0.4212 & 0.6463 & 0.7219 & 0.4354 & 0.2748 & 0.3221 & 0.7784 \\
    \checkmark &  & 0.4331 & \textbf{0.7834} & 0.6703 & 0.5594 & 0.2813 & 0.3741 & 0.8036 \\
     & \checkmark & 0.3845 & 0.7601 & 0.6575 & 0.4802 & 0.2452 & 0.3452 & 0.7966 \\
    \checkmark & \checkmark & \textbf{0.4548} & 0.7455 & \textbf{0.8705} & \textbf{0.5676} & \textbf{0.2982} & \textbf{0.3787} & \textbf{0.8620} \\
    \bottomrule[1pt]
    \end{tabular}
}
\label{tab:results4}
\end{table}

\begin{table}
    \caption{Quantitative results comparisons among different CLIP Text encoder and training strategies.}
    \begin{center}
        {
            \scalebox{0.62}{
                \begin{tabular}{llccccc}
                \toprule
                Text encoder & Parameters & Fine-tuned & IS $\uparrow$ & FID $\downarrow$ & CLIP Score $\uparrow$ & Zero-Shot classification OA $\uparrow$\\

                \midrule
                \multirow{2}{*}{ViT-B-32} & \multirow{2}{*}{63M} & \ding{55} & 15.73 & 83.21 & 19.93 & 55.46 \\
                                          &                      & \ding{51} & 16.22 & 73.94 & 19.96 & 63.26 \\
                \midrule
                \multirow{2}{*}{ViT-L-14} & \multirow{2}{*}{123M} & \ding{55} & 17.68 & 63.68 & 19.99 & 58.17 \\
                                          &                       & \ding{51} & \textbf{18.39} & \textbf{50.72} & \textbf{20.33} & \textbf{69.21} \\

                \bottomrule
                \end{tabular}
            }
        }
    \end{center}
    \label{tab:clip}
\end{table}

\subsection{Ablation Analysis}
We explore improvements in the structure of the multi-conditional control network and the method of injecting control information. We perform ablation experiments on CRS-Diff and its variants, analyzing the impact of replacing the backbone model (ReB) and the feature fusion approach (FF) on the generation quality and control effectiveness, respectively. We constructed a baseline based on the underlying Multi-ControlNet and executed the alteration approach sequentially, reporting the evaluation metrics of the different models, as shown in \cref{tab:results4}. The pre-training of the backbone model and the incorporation of the feature fusion approach significantly improve the generation effect, enabling the model to generate RS images with higher information density and realize the fusion of various control information types, further enhancing the model's control generation capability.

\begin{table}[h]
    \caption{Experimental results under different control conditions for CLIP models ViT-L-14 and ViT-B-32. The best experimental results are in \textbf{bold}, and the second-best results are \underline{underlined}.}
    \begin{center}
        {
            \scalebox{0.95}{
                \begin{tabular}{cccccc}
                \toprule
                CLIP Model & Text & Content image & IS $\uparrow$ & FID $\downarrow$ & CLIP Score $\uparrow$\\
                \midrule
                
                \multirow{3}{*}{ViT-B-32} 
                 & \ding{55} & \ding{51} & 10.17 & 172.15 & 0.7379 \\
                 & \ding{51} & \ding{55} & \underline{10.64} & 95.98 & 0.7532 \\
                 & \ding{51} & \ding{51} & 9.67 & \underline{72.19} & \underline{0.7671} \\
                \midrule
                \multirow{3}{*}{ViT-L-14} 
                 & \ding{55} & \ding{51} & \textbf{10.99} & 135.76 & 0.7311 \\
                 & \ding{51} & \ding{55} & 10.57 & 86.93 & 0.7573 \\
                 & \ding{51} & \ding{51} & 9.62 & \textbf{72.30} & \textbf{0.7674} \\

                \bottomrule
                \end{tabular}
            }
        }
    \end{center}
    \label{tab:condition}
\end{table}

\begin{table}[h]
    \caption{Quantitative results for different CLIP image encoders. The best experimental results are in \textbf{bold}, and the second-best results are \underline{underlined}.} 
    \begin{center}
        {
            \scalebox{1}{
                \begin{tabular}{lcccc}
                \toprule
                Image Encoder & Parameters (M) & IS $\uparrow$ & FID $\downarrow$ & CLIP Score $\uparrow$\\
                \midrule
                ResNet-50     & 38  & \textbf{10.78} & 183.33 & 0.7239 \\
                ResNet-101    & 56 & \underline{10.32} & 136.74 & 0.7383 \\
                \midrule
                ViT-B-32 & 87  & 10.15 & \underline{119.30} & \underline{0.7514} \\
                ViT-L-14 & 304 &  9.91 & \textbf{97.70} & \textbf{0.7528} \\
                \bottomrule
                \end{tabular}
            }
        }
    \end{center}
    \label{tab:clip_encoders}
\end{table}

\cref{tab:clip} presents a detailed ablation study to evaluate the impact of different versions of the CLIP model and the corresponding training strategy on the model generation capability in the single-text condition. We chose two versions of the CLIP text encoder for encoding the text condition in the text-to-image generation process, i.e., ViT-B-32 and ViT-L-14. We compared the effects of various parameter sizes and the impact of specific fine-tuning of the encoder on model performance. The results show that models with a greater number of parameters exhibit superior generation capabilities. Additionally, fine-tuning on RS images proves to be an effective method for enhancing performance.

We have conducted additional experiments to validate the positive impact of additional image control conditions on the quality of the generated images. We use a text encoder and an image encoder to process the text condition, the content image condition, and their combined conditions to generate the image, aiming to evaluate the quality of the generated images separately. The experimental results are shown in \cref{tab:condition}. As can be seen, this additional image condition information is beneficial to the quality of the generated images, reflecting on the FID and CLIP Score metrics. However, this image condition will reduce the diversity of the generated image, reflecting on the IS metric.

\cref{tab:clip_encoders} describes the impact of different versions of the CLIP image encoder on model generation capability guided by the content image condition. We compared the results of image generation using four different CLIP image encoders and found that encoders based on the ViT architecture consistently achieved the best results in terms of FID and CLIP scores, which are standard metrics for evaluating image quality. The ViT-based encoders demonstrated a significant performance advantage over those based on the ResNet architecture. This suggests that the CLIP ViT-L-14 model currently in use possesses superior feature extraction capabilities.

\begin{table}
    \caption{Experimental results comparison of road detection in the setting of different training datasets. Both the real and synthetic datasets consist of 1000 images. }
    \begin{center}
        {
            \resizebox{\linewidth}{!}{
                \begin{tabular}{lcccc}
                \toprule
                Train set & Road IOU $(\%)$ & Precision $(\%)$ & Recall $(\%)$ & F1 score $(\%)$\\
                \midrule
                Real & 52.84 & 59.26 & 85.51 & 69.49 \\
                Synthetic & 50.67 & 57.07 & 84.24 & 67.54 \\
                \midrule
                Real + Synthetic & \textbf{55.27} & \textbf{63.16} & \textbf{89.45} & \textbf{72.31} \\
                \bottomrule
                \end{tabular}
            }
        }
    \end{center}
    \label{tab:result4}
\end{table}

\begin{figure}
	\centering
	\includegraphics[width=0.9\linewidth]{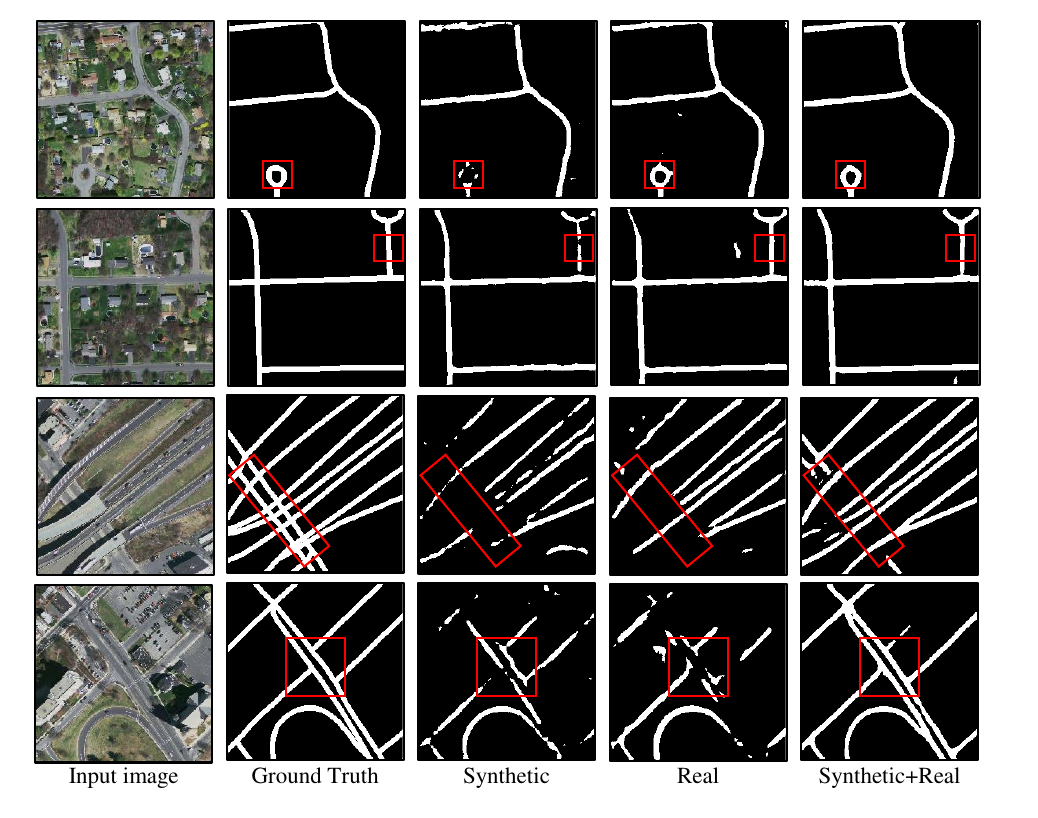}
	\caption{Visualisation comparison of SGCN model for road detection task under different settings of training datasets. }\label{fig:road}
\end{figure}

\subsection{Application for downstream road detection task}
For the condition generation phase, we posit that the generated image should have a sufficiently high correlation with the control labels as conditions. We aim for the generated image to encapsulate as much information from the conditioned image as possible and to offer training data support for downstream tasks. Consequently, we incorporate experiments on the generated images to ascertain the pertinent performance of CRS-Diff. We consistently integrate synthetic dummy data into the training set of SGCN \cite{zhou2021split} for the road extraction task and assess it on the official test set. \cref{tab:result4} demonstrates the performance comparison of the SGCM method under different settings of training datasets. As can be seen, synthetic training datasets can obtain almost the same performance as the real training dataset, which means that our CRS-Diff can simulate real images. By adding the synthetic dataset to the real dataset, the detection performance can be further significantly improved, indicating that the generated RS images conditioned on the road can promote the downstream road detection task. Besides, \cref{fig:road} also visually compares the road extraction results under the three training conditions. The red boxes highlight areas with relatively large differences, indicating that the model trained with the augmented dataset (Real + Synthetic) performs better in terms of continuity and completeness when dealing with complex road networks. Overall, the combination of real and synthetic data results in a more robust and generalized model, capable of handling more diverse and complex road structures. This blend of data sources not only increases the amount of training data but also introduces a wider range of scenarios, results in improved performance for road detection task.


\section{Conclusion}
\label{sec:conclusion}
In this paper, we propose a new controllable RS generative model with diffusion models (CRS-Diff). Developed from the diffusion model framework, CRS-Diff enables high-quality RS image generation. By integrating an optimized multi-conditional control mechanism, CRS-Diff can effectively synthesize multidimensional information, including text, metadata and images, guiding precise image generation and yielding highly accurate and controllable remote sensing images. This significantly broadens the control spectrum of the generation model, enhancing its adaptability to more complex application scenarios. Additionally, a comprehensive evaluation of existing multi-conditional generation models confirms CRS-Diff's superior ability to generate remote sensing images under various conditions and its high controllability. This makes it suitable for a wide spectrum of use cases and enhances the performance of downstream tasks.

\bibliographystyle{IEEEtran}

\bibliography{mybibfile}

\begin{IEEEbiography}[{\includegraphics[width=1in,height=1.25in,clip,keepaspectratio]{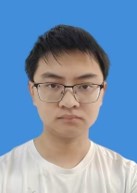}}]{Datao Tang}  received the B.E. degrees from Xi’an Jiaotong University, Xi’an, China, in 2023.
He is currently a postgraduate with the School of Computer Science and Technology, Xi’an Jiaotong University, Xi’an, China. His research interests include image processing and remote-sensing image generation.
\end{IEEEbiography}
		
\begin{IEEEbiography}[{\includegraphics[width=1in,height=1.25in,clip,keepaspectratio]{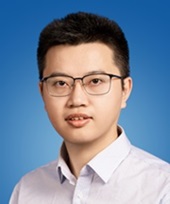}}] {Xiangyong Cao} (Member, IEEE) received the B.Sc. and Ph.D. degrees from Xi’an Jiaotong University, Xi’an, China, in 2012 and 2018, respectively. From 2016 to 2017, he was a Visiting Scholar with Columbia University, New York, NY, USA. He is an Associate Professor with the School of Computer Science and Technology, Xi’an Jiaotong University. His research interests include statistical modeling
and image processing.
\end{IEEEbiography}

\begin{IEEEbiography}[{\includegraphics[width=1in,height=1.25in,clip,keepaspectratio]{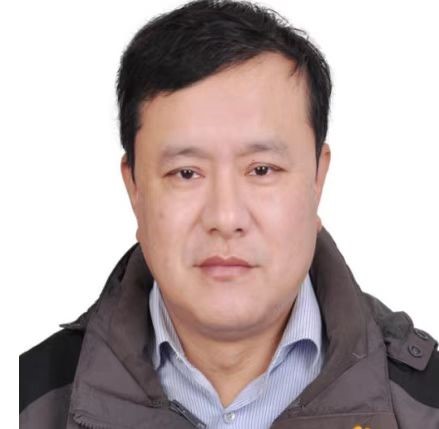}}] {Xingsong Hou} (Member, IEEE) received the Ph.D. degree from Xi’an Jiaotong University, China, in 2005. From October 2010 to October 2011, he was a Visiting Scholar at Columbia University, New York, NY, USA. He is currently a Professor with the School of Information and Communications Engineering, Xi’an Jiaotong University. He is also with the Key Laboratory for Intelligent Networks and Network Security, Ministry of Education. His research interests include video/image coding, wavelet analysis, sparse representation, compressive sensing, and radar signal processing.
\end{IEEEbiography}

\begin{IEEEbiography}[{\includegraphics[width=1in,height=1.25in,clip,keepaspectratio]{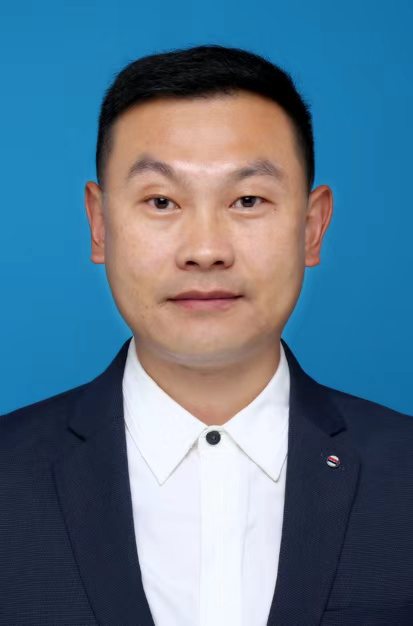}}]{Zhongyuan Jiang} received both B.S. and Ph.D. degrees from Beijing Jiaotong University in 2009 and 2013 respectively. Currently, he is a professor of School of Cyber Engineering, Xidian University, China. His research interests include privacy preserving, social computing, urban computing, and network functions virtualization.
\end{IEEEbiography}

\begin{IEEEbiography}[{\includegraphics[width=1in,height=1.25in,clip,keepaspectratio]{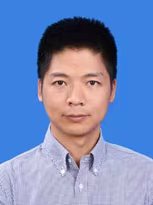}}]{Junmin Liu}
      received the Ph.D. degree in Mathematics from Xi’an Jiaotong University, Xi’an, China, in 2013. \par
	From 2011 to 2012, he has served as a Research Assistant with the Department of Geography and Resource Management, The Chinese University of Hong Kong, Hong Kong, China.
	From 2014 to 2017, he worked as a Visiting Scholar at the University of Maryland, College Park, USA. Currentlyk, he is a full Professor with the School of Mathematics and Statistics, Xi’an
	Jiaotong University, Xi’an, China. His main research interests include data mining, image processing, deep learning, and so on. He has published over 60+ research papers in international conferences and journals.
\end{IEEEbiography}

\begin{IEEEbiography}[{\includegraphics[width=1in,height=1.25in,clip,keepaspectratio]{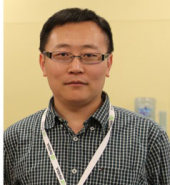}}]{Deyu Meng} (Member, IEEE) received the B.Sc., M.Sc., and Ph.D. degrees from Xi’an Jiaotong University, Xi’an, China, in 2001, 2004, and 2008, respectively.
From 2012 to 2014, he took his two-year sabbatical leave at Carnegie Mellon University, Pittsburgh, PA, USA. He is a Professor with the School of
Mathematics and Statistics, Xi’an Jiaotong University, and an Adjunct Professor with the Faculty of Information Technology, Macau University of Science and Technology, Taipa, Macau, China. His research interests include model-based deep learning, variational networks, and meta learning.
\end{IEEEbiography}

\end{document}